\documentclass[letterpaper, 10 pt, conference]{ieeeconf}  % Comment this line out if you need a4paper

\usepackage{amssymb}
\usepackage{amsmath}
\usepackage{graphicx}
\usepackage{color}
\usepackage{multirow}

\IEEEoverridecommandlockouts                              % This command is only needed if 
\title{\LARGE \bf
Reconstructed Student-Teacher and Discriminative Networks \\for Anomaly Detection 
}

\author{Shinji Yamada$^{1}$, Satoshi Kamiya$^{2}$ and Kazuhiro Hotta$^{3}$% <-this % stops a space
\thanks{$^{1}$Meijo University, 1-501 Shiogamaguchi, Tempaku-ku, Nagoya 468-8502, Japan,
        {\tt\small  160442146@ccalumni.meijo-u.ac.jp}}%
\thanks{$^{2}$Meijo University, 1-501 Shiogamaguchi, Tempaku-ku, Nagoya 468-8502, Japan,
        {\tt\small  180442042@ccalumni.meijo-u.ac.jp}}%
\thanks{$^{3}$Meijo University, 1-501 Shiogamaguchi, Tempaku-ku, Nagoya 468-8502, Japan,
        {\tt\small  kazuhotta@meijo-u.ac.jp}}%\textbf{}
}

\begin{document}

\maketitle
\thispagestyle{empty}
\pagestyle{empty}

% Resnet,ResNet,ResNeTを統一しResNetに変更しました
%%%%%%%%%%%%%%%%%%%%%%%%%%%%%%%%%%%%%%%%%%%%%%%%%%%%%%%%%%%%%%%%%%%%%%%%%%%%%%%%
% normal imagesを早めに定義しといたほうが良いとの意見です
% - I would recommend defining ‘normal images’ at least once early in the paper so that it’s % not vague to an uninformed reader. Just mentioning that you refer to images that do not
% contain anomalies as ‘normal images’ should suffice.

\begin{abstract}
Anomaly detection is an important problem in computer vision; however, the scarcity of 
 anomalous samples makes this task difficult. Thus, recent anomaly detection methods have used only "normal images" with no abnormal areas for training. In this work, a powerful anomaly detection method is proposed based on student-teacher feature pyramid matching (STPM), which consists of a student and teacher network. Generative models are another approach to anomaly detection. They reconstruct normal images from an input and compute the difference between the predicted normal and the input. Unfortunately, STPM does not have the ability to generate normal images. To improve the accuracy of STPM, this work uses a student network, as in generative models, to reconstruct normal features. This improves the accuracy; however, the anomaly maps for normal images are not clean because STPM does not use anomaly images for training, which decreases the accuracy of the image-level anomaly detection. To further improve accuracy, a discriminative network trained with pseudo-anomalies from anomaly maps is used in our method, which consists of two pairs of student-teacher networks and a discriminative network. The method displayed high accuracy on the MVTec anomaly detection dataset.
\end{abstract}

%%%%%%%%%%%%%%%%%%%%%%%%%%%%%%%%%%%%%%%%%%%%%%%%%%%%%%%%%%%%%%%%%%%%%%%%%%%%%%%%
\section{INTRODUCTION}

Anomaly detection is the identification of samples that deviate from normal patterns in the data. In recent years, convolutional neural networks (CNNs) have been applied to anomaly detection for the visual inspection of industrial products\cite{c2, c3, c4}, surveillance of locate intruders\cite{c5,c6,c7}, and pathological diagnosis of medical images\cite{c8, c9, c10, c11, c12}. In the past, image-level anomaly detection methods that classify each image as an anomaly have been studied, but image-level anomaly detection methods are not suitable for the inspection of products that require anomaly detection at the pixel level. Famous image-level anomaly detection methods rely on generative models to accurately reconstruct normal images. The generative model, based on the generative adversarial network (GAN)\cite{c11} and autoencoders \cite{c13}, can be trained using only normal images which allow the reconstruction of normal regions although anomalous regions cannot be reconstructed because the model is trained using only normal images. Anomaly detection using generative models identifies anomalous regions by visualizing regions with poor reconstruction capabilities. However, it is difficult for these methods to identify anomalous regions unless the generated normal regions can be reconstructed with high accuracy\cite{c15,c13}.

\begin{figure}[t]
\begin{center}

\includegraphics[scale=0.64]{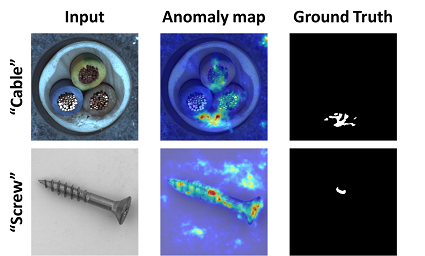}
\end{center}
\caption{The problem of STPM-based methods. The anomaly map has large values for both abnormal and normal regions.}
\label{f2}
\end{figure}

The use of pretrained models has been proposed as a new approach for anomaly detection, such as STPM\cite{c14} which uses a student-teacher network. In STPM, the teacher network is the pretrained ResNet18, and the student network is the untrained ResNet18. Only normal images are used for STPM training. The student network learns, thereby making feature maps similar to those in the teacher network. Because training is performed only on normal images, the student network can only output the features of normal regions. In contrast, the teacher network is a model pre-trained on ImageNet, so it can represent features of abnormal regions well. The difference in the feature representation between the student and teacher networks is the anomaly region.

In student ResNet18, STPM detects anomalous regions using anomaly maps at three different resolutions. However, STPMs have several limitations. STPM uses the difference between feature maps at three different resolutions in student and teacher networks. In the validation phase, the final anomaly map is calculated by multiplying the three maps. When all three anomaly maps detect the same abnormal area, highly accurate abnormality detection is possible. However, if one of the three maps fails to detect an abnormal region, the anomaly region cannot be detected because the three maps are multiplied.

Therefore, each of the three anomaly maps must be improved. Observing that STPM does not have the ability to reconstruct normal images similar to generative models which are effective in anomaly detection, we introduced a new student network that has the ability to reconstruct normal features. Conventional STPM uses ResNet18 as the teacher network. If the same teacher network is used for a new student, it is possible that similar anomaly maps can be obtained. To obtain anomaly maps from different viewpoints, the pre-trained ResNet50 was used as the teacher network for the student network for reconstruction. However, because the architecture of the student network is ResNet18, knowledge distillation for student-teacher pairs with different structures is more challenging than learning student-teacher pairs as in STPM with the same structure. Therefore, in the proposed method, to proceed with the learning, an attention mechanism is used with the structure to propagate some features of the teacher network to the student network. Because only normal images are used for training in the proposed method, the attention mechanism reconstructs normal regions with higher accuracy. Knowledge distillation is ensured by passing hints through the attention mechanism.

\begin{figure}[t]
\begin{center}
\vspace{0.2cm}
\includegraphics[scale=0.64]{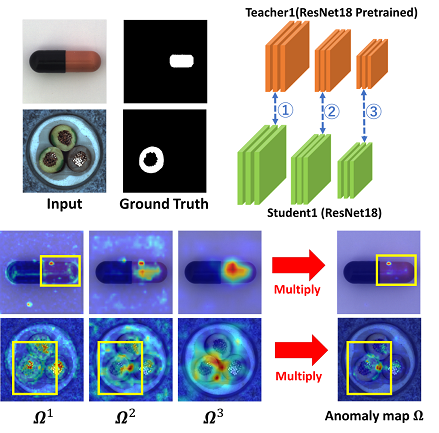}
\end{center}

\caption{STPM and its problems. STPM detects anomalies by multiplying the anomaly maps at three different resolutions. However, if one of the three anomaly maps fails to detect an anomaly location, the anomaly detection fails. The bottom images show examples of failures.}
\label{f1}
\end{figure}

Although the use of a student-teacher pair for reconstruction improves the accuracy of anomaly detection, a discriminative network is added to further improve the accuracy of anomaly detection. Because anomaly detection with knowledge distillation uses only normal images for training, it cannot discriminate whether a product is an anomaly. Thus, there are cases in which anomaly maps have large values for the normal regions, as shown in Figure \ref{f2}. Anomaly maps can detect abnormal areas, but they also detect normal areas as anomalies. This decreases the accuracy of anomaly detection.

Therefore, the anomaly maps obtained by our STPM, with two pairs of student-teacher networks, are input into a discriminative network, to reconsider the anomaly map when pseudo-abnormal images are also fed into our STPM.
The discriminative network thus learns to produce a more accurate anomaly map. The accuracy of anomaly detection is improved by multiplying the anomaly map obtained from the discriminant network by that obtained from the STPM with two pairs of networks. 

The proposed method was evaluated on an MVTec anomaly detection dataset\cite{c2}. The pixel-level AUC of the proposed method outperformed that of the original STPM in many categories. The image-level AUC of the proposed method was also better than that of the original STPM. In addition, the effectiveness of different teacher networks and attention mechanisms was demonstrated in ablation studies.

The rest of this paper is organized as follows. In Section II, related works are described. In Section III, the proposed method is explained in detail. In Section IV, the experimental results are presented. Finally, the study is concluded in Section V.

\begin{figure}[t]
\begin{center}
\vspace{0.2cm}
\includegraphics[scale=0.59]{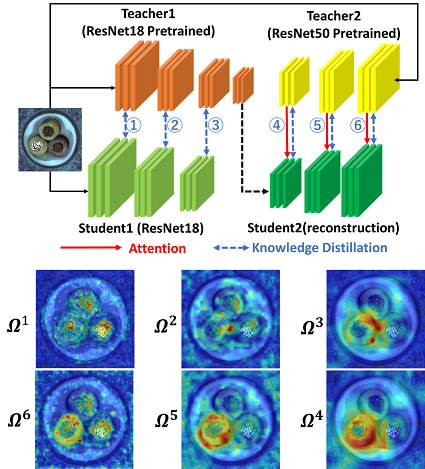}
\end{center}
\caption{Overview of STPM with two pairs of student-teacher networks. 
}
\label{f3}
\end{figure}

\section{Related Works}

\subsection{Anomaly Detection}

There are two types of anomaly detection: image-level and pixel-level. The goal of image-level anomaly detection is to classify anomalous samples correctly. There are three approaches for image-level anomaly detection: generative models, the distribution of data such as feature spaces, and classification. The generative model detects anomalies based on the degree of reconstruction loss\cite{c16, c17, c18}. The distribution-based method considers samples that deviate from the normal data distribution as abnormal. When generating a probability distribution of only normal products, the probability density of abnormal products is low\cite{c19, c20}, allowing the classification of abnormal data. A method based on classification is a method for anomaly detection that combines geometric transformation and classification\cite{c21}. It detects anomalous samples based on the idea that classification accuracy for unknown anomalous data is poor. However, these methods are unsuitable for anomaly detection at the pixel level.

Pixel-level anomaly detection is a method used for the detection of anomalies at each pixel. The difficulty of detecting anomalies at the pixel level is higher than that of detecting anomalies at the image level because the number of inspection targets increases. The mainstream method for detecting anomalies at the pixel level is based on generative models\cite{c13, c10, c11, c22}, such as GAN and autoencoders. In recent years, methods that combine GAN and autoencoders have been studied\cite{c23, c24, c25}. The generative model must reconstruct normal images with high accuracy; otherwise, the accuracy of anomaly detection decreases.

SPADE \cite{c26} was proposed as a new approach for pixel-level anomaly detection. SPADE compares the features of normal and abnormal images based on a pre-trained model, and k-means clustering was used to detect abnormalities. In addition to the pre-trained model, uninformed students \cite{c1} using knowledge distillation with a student-teacher structure have also been proposed. Uninformed students train the student network using only normal data. Anomaly maps are then calculated from two perspectives: (i) the difference between the outputs of student and teacher networks, (ii) the uncertainty of multiple student networks. STPM\cite{c14} was proposed as a method for developing both SPADE and uninformed students. In STPM, because the training is performed only on normal data, the student network represents only the features of the normal region. In contrast, the teacher network is pre-trained on ImageNet, so it can represent the features of anomalous regions. The difference in the feature representation between the student and teacher networks is the anomaly region. STPM uses knowledge distillation for feature maps at three different resolutions and outputs three anomaly maps. The final anomaly map is  computed by multiplying the three anomaly maps. However, if one of the three anomaly maps is unable to detect an anomalous region, the final anomaly map cannot detect the anomalous region, as shown in Figure \ref{f1}. To address this problem, each anomaly map is improved by using a new student network for reconstruction.

\subsection{Attention Mechanism}

Various attention mechanisms have been proposed for use in image recognition. The residual attention network \cite{c27} solved the vanishing gradient problem using a structure similar to that of a residual block. Squeeze-and-excitation network (SENet)\cite{c28} introduced an attention mechanism that emphasizes important channels in feature maps. A transformer \cite{c29} was proposed for language translation using only the attention mechanism. Several image recognition methods using self-attention, a type of transformer, have also been proposed\cite{c32, c33, c34}. An attention branch network\cite{c35} proposes an attention map for classification by aggregating multiple feature maps. An attention map can be used to visualize the basis of decisions.

In our method, the purpose of the attention mechanism is to leak the features of the teacher network to the student network to effectively reconstruct the features of normal data. Because anomalies must be detected at the pixel level, an attention mechanism that can emphasize and suppress pixels is more suitable than an attention mechanism that emphasizes channels such as SENet. If almost all features in the teacher network are leaked into the student network, there will be no difference between the student and teacher networks. Therefore, by aggregating the features in the teacher network into one channel, pixels can be emphasized and suppressed without providing the entire information. Therefore, we used an attention map generated from the teacher network. Because only normal data are used for training the student network, the attention mechanism is used to reconstruct the normal regions.

\section{Proposed Method}

The details of the proposed method are described in this section. The proposed method makes four contributions: the student network for reconstruction, attention from the teacher network to the student network, different teacher networks from the original STPM, and the discriminative network.

In Section III.A, the original STPM and its problems, in Section III.B, the student network for reconstruction and its teacher network, in Section III.C, the attention mechanism from the teacher to the student network, and in Section III.D, a discriminative network that revisits anomaly maps obtained by STPM with two pairs of student-teacher networks are described.

\subsection{Problem of STPM}

STPM is an effective anomaly detection method. STPM multiplies the three anomaly maps at different resolutions. The accuracy of the anomaly detection was improved by multiplying the three anomaly maps. However, if one of the three anomaly maps is unable to detect the anomalous region, the anomalous region cannot be detected after multiplying the three maps. Figure \ref{f1} shows an example in which STPM failed to detect anomalous regions  because of such a problem. Multiplying the three anomaly maps improved the accuracy of anomaly detection with respect to the detection accuracy at each resolution, but there were cases of failure after multiplication. To address this problem, it was necessary to improve each anomaly map at three different resolutions. As shown in the network structure of STPM in Figure \ref{f1}, STPM does not use feature reconstruction such as generative models. Adding reconstruction abilities, such as generative models, to new student networks was considered, to improve the accuracy of anomaly detection. The proposed network detects anomalies using two pairs of student-teacher networks.

\begin{figure}[t]
\begin{center}
\vspace{0.2cm}
\includegraphics[scale=0.70]{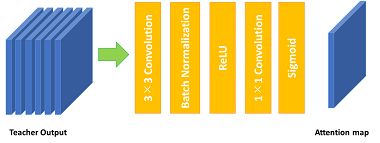}
\end{center}
\caption{Attention mechanism in the proposed method. 
The student network can emphasize the important pixels for reconstruction.}
\label{f4}
\end{figure}

\begin{figure}[t]
\begin{center}
\includegraphics[scale=0.64]{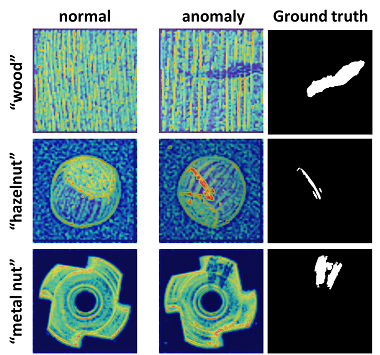}
\end{center}
\caption{Visualization of attention map. }
\label{f5}
\end{figure}

\begin{figure*}[t]
\begin{center}
\vspace{0.2cm}
\includegraphics[scale=0.64]{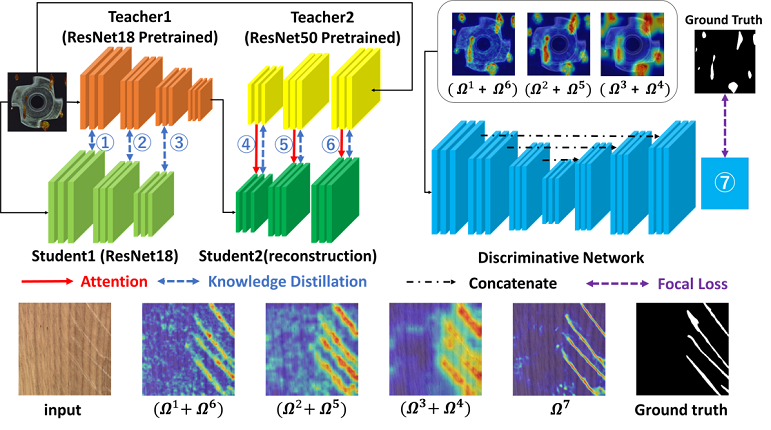}
\end{center}
\caption{Overview of the proposed network. The discriminant network uses pseudo-anomalies to reconsider the anomaly map obtained from STPM with two pairs of student-teacher networks. In the test phase, the anomaly map from the STPM is multiplied by that from the discriminant network.}
\label{f6}
\end{figure*}

\subsection{Student Network for Reconstruction} 

In this work, a new student network is introduced to reconstruct normal features. There are three main contributions in the new student network. The first is the reconstruction of normal features. Second, a teacher network different from that of the original STPM is used for the new student network. The third is the attention mechanism from the teacher to student network, as explained in the next section. 

Figure \ref{f3} presents an overview of the proposed method. The input for the new student network is the feature map at the lowest resolution in the pre-trained ResNet18, whereby the student network reconstructs the features of normal data. The structure of the new student network is similar to that of ResNet18. Therefore, if the same teacher network is used for student2 as in the original STPM, the same mistakes as in the original STPM will be made. Thus, ResNet50 pre-trained on ImageNet was used as the teacher network for student2 because ResNet50 has features and structures that are different from ResNet18, which is teacher1. With a perspective different from student1, such features will improve anomaly detection accuracy.

The equation for knowledge distillation is the same as that for conventional STPM. A normal image is defined as $I_k\in\mathbb{R} ^{w \times h \times c} $, where $h$ is the height, $w$ is the width, and $c$ is the number of channels. The features in the teacher and student networks are normalized along the channel dimensions as,
\begin{eqnarray}
\hat{F_t^l}{(I_k)}_{ij} = \frac{F^l_t(I_k)_{ij}}{ { || F^l_t(I_k)_{ij} || }_{\ell_2}^2 },\quad
\hat{F_s^l}{(I_k)}_{ij} = \frac{F^l_s(I_k)_{ij}}{ { || F^l_s(I_k)_{ij} || }_{\ell_2}^2 }
\end{eqnarray}
where $F_{t}^{l}$ and $F_{s}^{l}$ represent the features of the teacher and student networks, respectively; $l$ represents the resolution of the feature map; and $(i,j)$ is the position. The student network learns to obtain the same normalized output. The loss function is defined as follows:
\begin{eqnarray}
{\ell}^l(I_k)_{ij} =  \frac{1}{2} { || \hat{F_t^l}{(I_k)}_{ij} - \hat{F_s^l}{(I_k)}_{ij} || }_{\ell_2}^2
\end{eqnarray}

Equation 2 represents the pixel loss. For the entire image, the loss function is defined as,
\begin{eqnarray}
{\ell}^l(I_k) = \frac{1}{{w_l}{h_l}} \sum_{i=1}^{w_l} \sum_{j=1}^{h_l} {\ell^l}(I_k)_{ij}
\end{eqnarray}

The same loss function is used for feature maps at three resolutions in the two-student networks. The numbers in Fig. 3 indicate the locations where the loss function was applied. The final loss function is the sum of the losses, defined as,
\begin{eqnarray}
{\ell}(I_k) = \sum_{l=1}^{L} {{\ell}^l(I_k)}
\end{eqnarray}
%Ω1(J)などの説明がないという指摘がありました。
% How the anomaly map, $\Omega^{i}(J)$, is computed is never specified in the paper.  This should be included.

where L is the number of knowledge distillations. In this study, L = 6 because the loss was calculated at six locations, as shown in Figure \ref{f3}. In the test phase, six anomaly maps $\{\Omega^{1}(J),...,\Omega^{6}(J)\}$ were obtained using student1 and student2. As shown in Figure \ref{f3}, the anomaly maps at the same resolution are summed, while the anomaly maps at different resolutions are multiplied to generate the final anomaly map. When multiplying the three anomaly maps, the size of each anomaly map is made the same using bilinear interpolation. Anomaly detection is performed using the final anomaly map.

\begin{equation}
\begin{split}
\Omega(J) = {\{ \Omega^{1}(J) + \Omega^{6}(J) \}} {\odot} { \{ \Omega^{2}(J) + \Omega^{5}(J) \}}\\
{\odot} {\{ \Omega^{3}(J) + \Omega^{4}(J) \}}
\end{split}
\end{equation}
where $J\in \mathbb{R}^{w \times h \times c}$ denotes the test image. ${\odot}$ is the element-wise multiplication operation. The input for student2 is the feature maps with a size of 1/32 of the input image, and student2 reconstructs the feature maps up to 1/4 of the size of the input image. Anomaly maps from $\Omega^{1}$ to $\Omega^{3}$ in Figure \ref{f3} were obtained from the original STPM. Anomaly maps from $\Omega^{4}$ to $\Omega^{6}$ obtained from the reconstruction network are also shown. Anomaly map $\Omega^{4}$ has the lowest resolution. When the anomaly maps of the original STPM are compared with those of the reconstructed network, the accuracy of the reconstructed network is better than that of the original STPM. In particular, when the anomaly maps $\Omega^{1}$ are compared with $\Omega^{6}$ at the highest resolution, it is observed that the anomaly map $\Omega^{6}$ of the reconstructed network was able to detect anomalies that were not detected by the original STPM.

\subsection{Attention Mechanism for feature propagation} 

The third contribution of the proposed method is the attention mechanism from the teacher network to a new student network for reconstruction. The teacher network consisted of 50 layers, whereas the student network consisted of 18 layers. Knowledge distillation between networks with different structures is more difficult to learn than those between networks with the same structure\cite{c37}. It is likely that learning will not proceed as intended. Thus, an attention mechanism was used to ensure that learning proceeded well. Figure \ref{f4} shows the structure of the attention mechanism. The attention mechanism receives the feature maps in the teacher network and aggregates them into one channel. The attention mechanism was also used to teach hints to the student network to facilitate learning. Because only normal data are used for training, with the attention mechanism, the features in the student network for the normal region become more similar to the features in the teacher network. If most of the features in the teacher network are leaked into the student network, there will be no difference between the student and teacher networks. Thus, an attention map with only one channel was used to limit the amount of information provided. Considering an attention mechanism that emphasizes the channel would not be effective in detecting anomalies because it would not be able to learn important locations for pixel-level anomaly detection. Thus, considering a single attention map would be better because it would emphasize important pixels. The computational graph between the attention mechanism and the teacher network is not connected; therefore, the teacher network is not updated. Because only normal products are used as training data, the attention mechanism learns to successfully reconstruct the normal regions.

Figure \ref{f5} shows examples of attention maps. The attention maps at the highest resolution, which are feature map 6, are displayed in Figure \ref{f3}. Figure \ref{f5} shows the attention maps of the normal and abnormal data with three categories along with the ground truth labels corresponding to the abnormal regions. Attention maps differ between the normal and abnormal areas. The network was trained using only the normal data. Therefore, it behaves differently in normal regions when the input is an abnormal region that has never been seen during training. The attention mechanism was also introduced into the reconstruction network in which knowledge distillation is performed. The usage of the attention mechanism encourages the student network to represent the features in normal regions.

\subsection{Discriminative Network}
% studentとteacherネットワークはモデルのアーキテクチャが記述されているが識別ネットワークだけ記述されていないのがおかしいとの意見がありました。
% The discriminator network architecture is not defined anywhere in the paper. I would recommend adding information about it.

In this study, the anomaly detection accuracy of the conventional STPM was improved by adding a student network for reconstruction. The STPM with two pairs of student-teacher networks, uses only normal images for training. An anomaly map is calculated only by taking the difference in features between the student-teacher networks, and there are cases where the anomaly map has large values for normal regions. To address this problem, a discriminative network with U-Net was used. In an ideal anomaly map, the values for the normal regions are zero and only the abnormal regions have large values. Thus, a discriminative network was trained to learn pseudo-anomalies to make the anomaly map  closer to the ideal anomaly map. 

Figure \ref{f6} shows an overview of the proposed method with a discriminative network. Pseudo-anomalies are input into the trained STPM with two pairs of student-teacher networks and three anomaly maps are obtained. The anomaly maps at three resolutions, $\{ \Omega^{1}(I_a) + \Omega^{6}(I_a) \}$, $\{ \Omega^{2}(I_a) + \Omega^{5}(I_a) \}$, and $\{ \Omega^{3}(I_a) + \Omega^{4}(I_a) \}$, are fed into the discriminant network. Note that $I_a \in\mathbb{R} ^{w \times h \times c} $ is a pseudo-anomalous image. The discriminative network identifies pseudo-anomalies from the input anomaly maps. The focal loss used in training the difficult samples is defined as,
\begin{eqnarray}
L_{seg}(P, T) = \frac{1}{{h}{w}} \sum_{i=1}^{h} \sum_{j=1}^{w} {-T_{ij}(1-{P}_{ij})^{\gamma}\log {{P}_{ij}}}
\end{eqnarray}
where $P$ is the anomaly map for the pseudo-anomaly predicted by the discriminative network; $T$ is the ground truth; and $h$ and $w$ denote the height and width of the input image, respectively. $\gamma$ was set to two. Because the discriminative network learns to discriminate whether the regions are anomalous, anomaly values for normal regions are closer to zero, and anomaly regions have values closer to one. The anomaly map from our STPM was combined with that of the discriminative network to achieve more accurate anomaly detection. The final anomaly map was calculated as follows:
\begin{equation}
\begin{split}
\Omega(J) = {\{ \Omega^{1}(J) + \Omega^{6}(J) \}} {\odot} { \{ \Omega^{2}(J) + \Omega^{5}(J) \}}\\
{\odot} {\{ \Omega^{3}(J) + \Omega^{4}(J) \}} {\odot} {\{ \Omega^{7}(J)\}}
\end{split}
\end{equation}
where $\Omega^{7}$ is the anomaly map of the discriminant network and $J$ is a test image. ${\odot}$ is the element-wise multiplication operation.

The pseudo-anomalies proposed in DRAEM\cite{c38} were used to train the discriminative network. DRAEM can generate various types of pseudo-anomalies using Perlin noise\cite{c39} as well as parameters that determine the concentration of anomalous regions. In this study, DRAEM pseudo-anomalies were used to prevent overfitting of pseudo-anomalies.

\section{Experiments}

In this section, the experimental results are presented. In Section IV.A, the datasets and experimental conditions are described. In Section IV.B, the experimental results for the MVTec dataset are presented. In Section IV.C,  the results of the ablation study are presented.

\subsection{Datasets} 

The MVTec anomaly datasets\cite{c2} was used for the evaluation. The MVTecAD includes 15 categories of industrial products. The training data consisted of only the normal products. The image size was different for each category, and all images were resized to 256 × 256 pixels. The STPM was trained with two pairs of student-teacher networks and separately with the discriminative network. The numbers of epochs for training the STPM and discriminative network were 100 and 300, respectively. The SGD optimizer was used with a momentum of 0.9. The learning rate was set to 0.4, batch size to 32, and weight decay to $1 \times 10^{-4}$. The optimization method was Adam in the discriminative network, and the learning rate was set to 0.0001.

The pixel-level and image-level areas under the ROC curve (AUC) were used as the evaluation measures. Per-region-overlap (PRO)\cite{c1} was also used to evaluate the anomaly maps.

\begin{figure}[t]
\begin{center}
\vspace{0.2cm}
\includegraphics[scale=0.59]{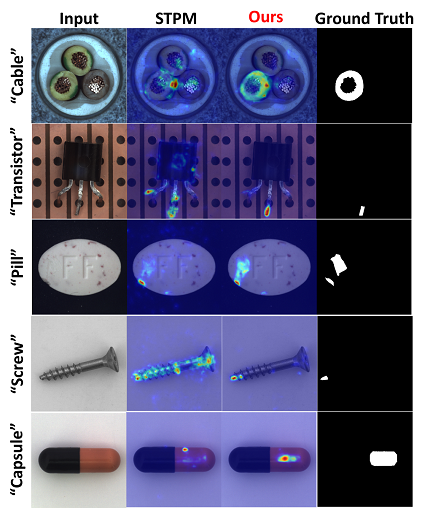}
\end{center}
\caption{Examples of anomaly map.}
\label{f7}
\end{figure}

\subsection{Results on Anomaly detection}
% TABLE1のOurs(Discriminative)が何かわからないとのことでした。
% What exactly is being tested is sometimes confusing.  For example, when discussing the results the paper says "With our (discriminative) method only the results using the discriminative network are obtained." Does this mean that only the discriminative network that produces anomaly map Omega-7 was used?  Or does this mean the expression in equation 7 is used to produce the anomaly map?

% 変更前：With our (discriminative) method only the results using the discriminative network are obtained.

First, pixel-level AUC and PRO metrics are described. The pixel-level AUC and PRO metrics indicate whether the anomaly map correctly detects the anomalous area. Table \ref{t1} displays the accuracy. Our (Student-Tescher) method in Table \ref{t1} shows the results using only the network in Figure \ref{f3}. Our (Discriminative) method shows the accuracy when only the output $\Omega^{7}$ of the discriminative network is used. It is observed from the table that the reconstruction network when combined with the conventional STPM improves the accuracy, and the discriminative network alone outperforms the conventional STPM. The anomaly detection accuracy of the proposed method using both the STPM with two pairs of student-teacher networks and the discriminative network, as shown in Figure \ref{f6}, provided further improvement. The pixel-level AUC and PRO scores of the proposed method outperformed the accuracies of the DRAEM using pseudo-anomalies and Padim using a pre-trained model. The proposed method can accurately detect anomalous regions because the reconstruction network improves the conventional STPM and the discriminative network assists the STPM using the reconstruction network. 

Next, the results of image-level AUC scores are presented. The image-level AUC is a score that indicates whether the input image can be correctly classified as abnormal. Table \ref{t2} lists the image-level AUCs. The accuracy of anomaly detection using the proposed methods outperformed conventional STPM. In the Table, the anomaly classification accuracies of Padim, STPM, and our (student-teacher) method which learn only normal products are lower than those of DRAEM and our (discriminative) method trained using pseudo-anomalies. The accuracy of anomaly classification can be improved using pseudo-anomalies. The proposed method achieved the best accuracy by combining the reconstruction and discriminative networks.

The qualitative results are shown in Figure \ref{f7}. The anomaly maps of the proposed method displayed in Figure \ref{f7}, are significantly better than those of conventional STPM. Conventional STPM cannot accurately detect anomaly areas of "Cable", "Pill", and "Capsule". In contrast, the proposed method can accurately detect anomalous regions that STPM cannot. STPM detects the abnormal area of "Screw" and "Transistor,” but there are large anomaly values in normal areas. The proposed method can accurately detect these anomalies. These results demonstrate the effectiveness of the proposed method.

\begin{table*}
\begin{center}
\vspace{0.2cm}
\caption{Anomaly localization on the MVTec (Pixel level AUROC / PRO AUROC)}
\label{t1}
\scalebox{0.89}{
\begin{tabular}{cccccccc}
\hline
\multicolumn{1}{l}{}      & category   & PaDim\cite{c40}         & DRAEM      & STPM          & \begin{tabular}[c]{@{}c@{}}Ours\\      (Student-Teacher)\end{tabular} & \begin{tabular}[c]{@{}c@{}}Ours\\      (Discriminative)\end{tabular} & Ours          \\ \hline
\multirow{5}{*}{\rotatebox[origin=c]{90}{Textures}} & Carpet     & {0.991 / 0.962} & {0.955 /  }- & {0.988 / 0.958} & {0.988 / 0.952} & {0.985 / 0.952} & {\textbf{0.992} / \textbf{0.968}} \\
                          & Grid       & {0.973 / 0.946} & {\textbf{0.997} / -}  & {0.990 / 0.966} & {0.994 / 0.967} & {0.996 / 0.974} & {0.996 / \textbf{0.975}} \\
                          & Leather    & {0.992 / 0.978} & {0.986 / -}  & {0.993 / 0.980} & {0.984 / 0.951} & {\textbf{0.996} / \textbf{0.984}} & {\textbf{0.996} / \textbf{0.984}} \\
                          & Tile       & {0.941 / 0.860} & {\textbf{0.992} / -}  & {0.974 / 0.921} & {0.970 / 0.910} & {0.987 / 0.958} & {0.988 / \textbf{0.962}} \\
                          & Wood       & {0.949 / 0.911} & {0.964 / -}  & {0.972 / 0.936} & {0.970 / 0.942} & {0.978 / 0.960} & {\textbf{0.981} / \textbf{0.967}} \\ \hline
\multirow{10}{*}{\rotatebox[origin=c]{90}{Objects}} & Bottle     & {0.983 / 0.948} & {0.991 / -}  & {0.988 / 0.951} & {0.990 / 0.965} & {\textbf{0.993} / 0.962} & {\textbf{0.993} / \textbf{0.973}} \\
                          & Cable      & {0.967 / 0.888} & {0.943 / -}  & {0.955 / 0.877} & {0.973 / 0.924} & {0.979 / 0.926} & {\textbf{0.983} / \textbf{0.948}} \\
                          & Capsule    & {\textbf{0.985} / \textbf{0.935}} & {0.943 / -}  & {0.983 / 0.922} & {\textbf{0.985} / 0.917} & {0.972 / 0.874} & {\textbf{0.985} / 0.910} \\
                          & Hazel nut  & {0.982 / 0.926} & {\textbf{0.997} / -}  & {0.985 / 0.943} & {0.991 / 0.948} & {0.995 / 0.949} & {0.995 / \textbf{0.956}} \\
                          & Metal nut  & {0.972 / 0.856} & {\textbf{0.995} / -}  & {0.976 / 0.945} & {0.982 / 0.947} & {0.989 / 0.958} & {0.989 / \textbf{0.963}} \\
                          & Pill       & {0.957 / 0.927} & {0.976 / -}  & {0.978 / 0.965} & {0.972 / 0.965} & {\textbf{0.990} / 0.961} & {0.987 / \textbf{0.974}} \\
                          & Screw      & {0.985 / 0.944} & {0.976 / -}  & {0.983 / 0.930} & {\textbf{0.993} / \textbf{0.958}} & {0.989 / 0.935} & {\textbf{0.993} / 0.955} \\
                          & Toothbrush & {0.988 / \textbf{0.931}} & {0.981 / -}  & {0.989 / 0.922} & {0.989 / 0.905} & {\textbf{0.994} / 0.925} & {0.993 / 0.927} \\
                          & Transistor & {\textbf{0.975} / \textbf{0.845}} & {0.909 / -}  & {0.825 / 0.695} & {0.898 / 0.817} & {0.872 / 0.803} & {0.907 / 0.830} \\
                          & Zipper     & {0.985 / 0.959} & {0.988 / -}  & {0.985 / 0.952} & {0.985 / 0.955} & {0.988 / 0.957} & {\textbf{0.992} / \textbf{0.972}} \\ \hline
\multicolumn{1}{l}{}      & Mean       & {0.975 / 0.921} & {0.973 / -}  & {0.970 / 0.921} & {0.977 / 0.935} & {0.980 / 0.939} & {\textbf{0.985} / \textbf{0.951}} \\ \hline
\end{tabular}
}
\end{center}
\end{table*}

\begin{table*}
\begin{center}
\caption{Result of image level AUC on the MVTec}
\label{t2}
\scalebox{0.89}{
\begin{tabular}{cccccccc}
\hline
\multicolumn{1}{l}{}      & category   & PaDim\cite{c40}          & DRAEM          & STPM  & \begin{tabular}[c]{@{}c@{}}Ours\\      (Student-Teacher)\end{tabular} & \begin{tabular}[c]{@{}c@{}}Ours\\      (Discriminative)\end{tabular} & Ours           \\ \hline
\multirow{5}{*}{\rotatebox[origin=c]{90}{Textures}} & Carpet     & \textbf{0.998} & 0.970          & -     & 0.981                                                                 & 0.964                                                                & 0.987          \\
                          & Grid       & 0.967          & 0.999          & -     & 0.984                                                                 & \textbf{1.000}                                                       & \textbf{1.000} \\
                          & Leather    & \textbf{1.000} & \textbf{1.000} & -     & 0.998                                                                 & \textbf{1.000}                                                       & \textbf{1.000} \\
                          & Tile       & 0.981          & 0.996          & -     & 0.953                                                                 & 0.986                                                                & \textbf{0.999} \\
                          & Wood       & 0.992          & 0.991          & -     & 0.992                                                                 & 0.966                                                                & \textbf{0.993} \\ \hline
\multirow{10}{*}{\rotatebox[origin=c]{90}{Objects}} & Bottle     & 0.999          & 0.992          & -     & \textbf{1.000}                                                        & \textbf{1.000}                                                       & \textbf{1.000} \\
                          & Cable      & 0.927          & 0.918          & -     & 0.967                                                                 & 0.989                                                                & \textbf{0.996} \\
                          & Capsule    & 0.913          & \textbf{0.985} & -     & 0.873                                                                 & 0.945                                                                & 0.930          \\
                          & Hazel nut  & 0.920          & \textbf{1.000} & -     & \textbf{1.000}                                                        & 0.987                                                                & 0.998          \\
                          & Metal nut  & 0.987          & 0.987          & -     & \textbf{1.000}                                                        & 0.999                                                                & \textbf{1.000} \\
                          & Pill       & 0.933          & \textbf{0.989} & -     & 0.967                                                                 & 0.980                                                                & 0.981          \\
                          & Screw      & 0.858          & 0.939          & -     & 0.948                                                                 & 0.941                                                                & \textbf{0.968} \\
                          & Toothbrush & 0.961          & \textbf{1.000} & -     & 0.900                                                                 & 0.987                                                                & 0.979          \\
                          & Transistor & 0.974          & 0.931          & -     & 0.975                                                                 & 0.978                                                                & \textbf{0.983} \\
                          & Zipper     & 0.903          & \textbf{1.000} & -     & 0.898                                                                 & 0.998                                                                & 0.993          \\ \hline
\multicolumn{1}{l}{}      & Mean       & 0.955          & 0.980          & 0.955 & 0.962                                                                 & 0.981                                                                & \textbf{0.987} \\ \hline
\end{tabular}
}
\end{center}
\end{table*}

\begin{table}[t]
\begin{center}
\caption{Accuracy while changing teacher network for reconstruction student}
\label{t3}
\scalebox{1.0}{
\begin{tabular}{c|cc}
\hline
                & ResNet18 & ResNet50(Ours) \\ \hline
Image Level AUC & 0.982    & \textbf{0.987} \\
Pixel Level AUC & 0.979    & \textbf{0.985} \\
PRO             & 0.937    & \textbf{0.951} \\ \hline
\end{tabular}
}
\end{center}
\end{table}

\begin{table}[t]
\begin{center}
\caption{Accuracy with/without attention mechanism}
\label{t4}
\scalebox{0.93}{
\begin{tabular}{c|ccc}
\hline
                & Without Attention & All Students–Teachers & Ours           \\ \hline
Image Level AUC & 0.985             & 0.985               & \textbf{0.987} \\
Pixel Level AUC & 0.983             & 0.983               & \textbf{0.985} \\
PRO             & 0.944             & 0.946               & \textbf{0.951} \\ \hline
\end{tabular}
}
\end{center}
\end{table}

\subsection{Ablation study}

In the proposed method, student network1 and student network2 have different teacher networks, as shown in Figure \ref{f3} and 6, respectively. An attention mechanism was also used from the teacher network to the student network. First, the influence of changes in teacher networks is discussed. In this study, ResNet50 was used as a teacher for student reconstruction. The proposed method was compared with the method in which ResNet18 was used as a teacher. Note that all the methods use an attention mechanism from the teacher-to-student network. Table \ref{t3} lists the accuracy associated with the teacher network changes. Different teacher networks lead to a perspective different from that of the conventional method, improving anomaly detection accuracy.

Next, the influence of the attention mechanism is discussed. In this study, an attention mechanism was used only between Student2 and Teacher2. The proposed method is compared with the method without the attention mechanism and the method using attention for both Student1-Teacher1 and Student2-Teacher2. Table \ref{t4} shows that it is better to use the attention mechanism only between Student2 and Teacher2. We believe that the attention mechanism allows us to learn better. 
Attention between Student1 and Teacher1 did not improve the accuracy. Because Student1 and Teacher1 have the same structure, knowledge distillation is easy. Therefore, if the attention mechanism is used for student1 and teacher1, the representational ability of the student network is high, possibly leading to a small difference in features between students and teachers. The attention mechanism was appropriate for use in Student2 and Teacher2, providing the expected improvement.

\section{Conclusions}

In this study, the STPM anomaly detection method was improved using a new student network for reconstruction. The teacher network was modified for a new student network and the attention mechanism from teacher to student was used to ensure successful learning. To further improve the accuracy of anomaly detection, discriminative networks were used to reconsider the anomaly maps. By using pseudo-anomalies in training a discriminative network, a more accurate anomaly map was generated for normal regions. The proposed method, using both STPM with two pairs of student-teacher networks and a discriminative network, achieved high anomaly detection accuracy compared with conventional methods.

\addtolength{\textheight}{-12cm}   

%%%%%%%%%%%%%%%%%%%%%%%%%%%%%%%%%%%%%%%%%%%%%%%%%%%%%%%%%%%%%%%%%%%%%%%%%%%%%%%%

%\section*{ACKNOWLEDGMENT}

%%%%%%%%%%%%%%%%%%%%%%%%%%%%%%%%%%%%%%%%%%%%%%%%%%%%%%%%%%%%%%%%%%%%%%%%%%%%%%%%


\begin{thebibliography}{99}

\bibitem{c1} Bergmann, P., Fauser, M., Sattlegger, D., Steger, C., Uninformed students: Student-teacher anomaly detection with discriminative latent embeddings. In: Proceedings of the IEEE/CVF Conference on Computer Vision and Pattern Recognition, pp. 4183-4192, 2020.
\bibitem{c2} Bergmann, P., Fauser, M., Sattlegger, D., Steger, C., Mvtec-ad real real-world dataset for unsupervised anomaly detection. In: Proceedings of the IEEE/CVF Conference on Computer Vision and Pattern Recognition, pp. 9592-9600, 2019.
\bibitem{c3} Bergmann, P., Fauser, M., Sattlegger, D., Steger, C., Uninformed students: Student-teacher anomaly detection with discriminative latent embeddings. In: Proceedings of  the IEEE/CVF Conference on Computer Vision and Pattern Recognition, pp. 4183-4192, 2020.
\bibitem{c4} Napoletano, P., Piccoli, F., Schettini, R., Anomaly detection in nano brous materials by cnn-based self-similarity. Sensors18(1), 209, 2018.
\bibitem{c5}Abati, D., Porrello, A., Calderara, S., Cucchiara, R., Latent  space  autoregression for novelty detection. In: Proceedings of the IEEE/CVF Conference on Computer Vision and Pattern Recognition, pp. 481-490, 2019.
\bibitem{c6} Nguyen, P., Liu, T., Prasad, G., Han, B., Weakly supervised action localization by sparse temporal pooling network. In: Proceedings of the IEEE/CVF Conference on Computer Vision and Pattern Recognition, pp. 6752-6761, 2018.
\bibitem{c7} Roitberg, A., Al-Halah, Z., Stiefelhagen, R., Informed democracy: voting-based novelty detection for action recognition. arXiv preprint arXiv:1810.12819, 2018.
\bibitem{c8} Baur, C., Wiestler, B., Albarqouni, S., Navab, N., Deep autoencoding models for unsupervised anomaly segmentation in brain MR images. In: International MICCAI Brain lesion Workshop, pp. 161-169, 2018.
\bibitem{c9} Li, L., Xu, M., Wang, X., Jiang, L., Liu, H., Attention based glaucoma detection: a large-scale database and CNN model. In: Proceedings of the IEEE/CVF Conference on Computer Vision and Pattern Recognition, pp. 10571-10580, 2019.
\bibitem{c10} Schlegl, T., Seeböck, P., Waldstein, S.M., Langs, G., Schmidt-Erfurth, U., f-anoGAN: Fast unsupervised anomaly detection with generative adversarial networks. Medical Image Analysis, 54, 30-44, 2019.
\bibitem{c11} Schlegl, T., Seeböck, P., Waldstein, S.M., Schmidt-Erfurth, U., Langs, G., Unsupervised anomaly detection with generative adversarial networks to guide marker discovery. In: International Conference on Information Processing in Medical Imaging, pp. 146-157, 2017.
\bibitem{c12} Vasilev, A., Golkov, V., Meissner, M., Lipp, I., Sgarlata, E., Tomassini, V., Jones,D.K., Cremers, D., Q-space novelty detection with variational autoencoders. In: Computational Diffusion MRI, pp. 113-124, 2020.
\bibitem{c13} Bergmann, P., Löwe, S., Fauser,  M., Sattlegger, D., Steger, C., Improving unsupervised defect segmentation by applying structural similarity to autoencoders. arXiv preprint arXiv:1807.02011, 2018.
\bibitem{c14} Wang, G., Han, S., Ding, E., Huang, D., Student-teacher feature pyramid matching for unsupervised anomaly detection. arXiv preprint arXiv:2103.04257, 2021.
\bibitem{c15} Nalisnick, E., Matsukawa, A., Teh, Y.W., Gorur, D., Lakshminarayanan, B., Do deep generative models know what they don't know? arXiv Preprint arXiv:1810.09136, 2018.
\bibitem{c16} An, J., Cho, S., Variational autoencoder based anomaly detection using reconstruction probability. Special Lecture on IE 2(1), 1-18, 2015.
\bibitem{c17} Zhou, C., Paffenroth, R.C., Anomaly detection with robust deep autoencoders. In: Proceedings of the 23rd ACM SIGKDD International Conference on Knowledge Discovery and Data Mining, pp. 665-674, 2017.
\bibitem{c18} Zenati, H., Romain, M., Foo, C.S., Lecouat, B., Chandrasekhar, V., Adversarially learned anomaly detection. In:  2018 IEEE International Conference on Data Mining, pp. 727-736, 2018.
\bibitem{c19} Deecke, L., Vandermeulen, R., Ruff, L., Mandt, S., Kloft, M., Image anomaly detection with generative adversarial networks. In:  Joint European Conference on Machine Learning and Knowledge Discovery in Databases, pp. 3-17, 2018.
\bibitem{c20} Zong, B., Song, Q., Min, M.R., Cheng, W., Lumezanu, C., Cho, D., Chen, H., Deep autoencoding gaussian mixture model for unsupervised anomaly detection. In: International Conference on Learning Representations, 2018.
\bibitem{c21} Golan, I., El-Yaniv, R., Deep anomaly detection using geometric transformations. Advances in Neural Information Processing Systems, 31, 2018.
\bibitem{c22} Zavrtanik, V., Kristan, M., and Skočaj, D., Reconstruction by inpainting for visual anomaly detection. Pattern Recognition, 112, 107706, 2021.
\bibitem{c23} Akcay, S., Atapour-Abarghouei, A., Breckon, T.P., Ganomaly: Semi-supervised anomaly detection via adversarial training. In: Asian Conference on Computer Vision, pp. 622-637, 2018.
\bibitem{c24} Akcay, S., Atapour-Abarghouei, A., Breckon, T.P., Skip-ganomaly: Skip connected and adversarially trained encoder-decoder anomaly detection. In: 2019 International Joint Conference on Neural Networks, pp. 1-8, 2019.
\bibitem{c25} Tang, T.W., Kuo, W.H., Lan, J.H., Ding, C.F., Hsu, H., Young, H.T., Anomaly detection neural network with dual auto-encoders GAN and its industrial inspection applications. Sensors 20(12), 3336, 2020.
\bibitem{c26} Cohen, N., Hoshen, Y., Sub-image anomaly detection with deep pyramid correspondences. arXiv Preprint arXiv:2005.02357, 2020.
\bibitem{c27} Wang, F., Jiang, M., Qian C., Yang, S., Li, C., Zhang, H., Wang, X., Tang, X., Residual attention network for image classification. In: Proceedings of the IEEE/CVF Conference on Computer Vision and Pattern Recognition, pp. 3156-3164, 2017.
\bibitem{c28} Hu, J., Shen, L., Sun, G., Squeeze-and-excitation networks. In: Proceedings of the IEEE/CVF Conference on Computer Vision and Pattern Recognition, pp. 7132-7141, 2018.
\bibitem{c29} Vaswani, A., Shazeer, N., Parmar, N., Uszkoreit, J., Jones, L., Gomez, A.N., Kaiser, L., Polosukhin, I., Attention is all you need. In: Advances in Neural Information Processing Systems, pp. 5998-6008, 2017.
%\bibitem{c30} Singh, P., Mazumder, P., Namboodiri, V., Accuracy booster: Performance boosting using feature map re-calibration. In: Proceedings of the IEEE/CVF Winter Conference on Applications of Computer Vision, pp. 884-893, 2020.
%\bibitem{c31} Wang, Q., Wu, B., Zhu, P., Li, P., Zuo, W., Hu, Q., Eca-net: efficient channel attention for deep convolutional neural networks, In: Proceedings of the IEEE/CVF Conference on Computer Vision and Pattern Recognition, pp. 11531-11539, 2020.
\bibitem{c32} Fu, J., Liu, J., Tian, H., Li, Y., Bao, Y., Fang, Z., Lu, H., Dual attention network for scene segmentation. In: Proceedings of the IEEE/CVF Conference on Computer Vision and Pattern Recognition, pp. 3146-3154, 2019.
\bibitem{c33} Ramachandran, P., Parmar, N., Vaswani, A., Bello, I., Levskaya, A., Shlens, J., Stand-alone self-attention in vision models. arXiv preprint arXiv:1906.05909, 2019.
\bibitem{c34} Zhang, H., Goodfellow, I., Metaxas, D., Odena, A., Self-attention generative adversarial networks. In: International Conference on Machine Learning, pp.7354-7363, 2019.
\bibitem{c35} Fukui, H., Hirakawa, T., Yamashita, T., Fujiyoshi, H., Attention branch network : Learning of attention mechanism for visual explanation. In: Proceedings of the IEEE/CVF Conference on Computer Vision and Pattern Recognition, pp. 10705-10714, 2019.
%\bibitem{c36} Zhou, B., Khosla, A., Lapedriza, A., Oliva, A., Torralba, A., Learning deep features for discriminative localization. In: Proceedings of the IEEE/CVF Conference on Computer Vision and Pattern Recognition, pp. 2921-2929, 2016.
\bibitem{c37} Wang, L., Yoon, K.J., Knowledge distillation and student-teacher learning for visual intelligence: A review and new outlooks. IEEE Transactions on Pattern Analysis and Machine Intelligence, 2021.
\bibitem{c38} Zavrtanik, V., Kristan, M., and Skočaj, D., DRAEM-A discriminatively trained reconstruction embedding for surface anomaly detection. In Proceedings of the IEEE/CVF International Conference on Computer Vision, pp. 8330-8339, 2021.
\bibitem{c39} Perlin, K., An image synthesizer. ACM Siggraph Computer Graphics, 19(3), pp. 287-296, 1985.
\bibitem{c40} Defard, T., Setkov, A., Loesch, A., and Audigier, R., Padim: a patch distribution modeling framework for anomaly detection and localization. In International Conference on Pattern Recognition, pp. 475-489, 2021.

\end{thebibliography}
\end{document}